# On Modeling Profiles instead of Values

**Alon Orlitsky** **Narayana P. Santhanam** **Krishnamurthy Viswanathan** **Junan Zhang**
ECE Dept.
University of California, San Diego
La Jolla, CA 92093

## Abstract

We consider the problem of estimating the distribution underlying an observed sample of data. Instead of maximum likelihood, which maximizes the probability of the observed values, we propose a different estimate, the *high-profile* distribution, which maximizes the probability of the observed *profile*—the number of symbols appearing any given number of times. We determine the high-profile distribution of several data samples, establish some of its general properties, and show that when the number of distinct symbols observed is small compared to the data size, the high-profile and maximum-likelihood distributions are roughly the same, but when the number of symbols is large, the distributions differ, and high-profile better explains the data.

## 1 INTRODUCTION

The problem of identifying among a collection of distributions the one that best explains an observed sample of data arises in many applications. In several biological studies, the number and distribution of species populations needs to be estimated based on a modestly-sized sample. In network management, the distribution of message lengths and volume, or of requested routes, can help design protocols and buffer sizes that best handle the expected demand. And in computer security, knowledge of the distribution of the observed traffic can help determine when an unlikely, possibly alarming, event has occurred.

One of the most popular approaches to this distribution-estimation problem is *maximum-likelihood estimation* that selects the distribution maximizing the data's probability. For example, upon observing a sequence of 30 $a$'s and 70 $b$'s generated by an *i.i.d.* Bernoulli($p$) distribution, maximum likelihood would postulate that $p \stackrel{\text{def}}{=} p(a) = 0.3$, the value maximizing the sequence probability $p^{30}(1-p)^{70}$. This is a natural estimate that also coincides with the intuitive *empirical-frequency* estimation assigning to each symbol a probability corresponding to the fraction of times it appeared.

Maximum-likelihood performs well when the sample size is large compared to the source alphabet. However, when, as in the above applications, the sample size is comparable to or smaller than the size of the source's alphabet, maximum likelihood may fall short. Consider sampling the DNA sequences of $n$ individuals. Clearly, all $n$ sequences will be distinct, and maximum likelihood would hypothesize a uniform distribution over the set of $n$ observed sequences. However, this distribution does not capture the essential property that no two sequences are identical. A distribution over a much larger, essentially infinite, range of sequences, will better explain the data.

In this paper we consider a modification of maximum likelihood that coincides with it when the sample size is large relative to the number of possible outcomes, but better models the data for smaller sample sizes. It is based on the simple premise that the underlying distribution should be estimated based not on the specific values observed, but only on the number of values appearing any given number of times.

For example, suppose that the first three outputs of an unknown *i.i.d.* source are

@@#.

Maximum likelihood would hypothesize that the symbols are generated by a $(2/3, 1/3)$ distribution where $2/3$ is the probability of @ and $1/3$ is the probability of #, yielding maximum-likelihood probability of $4/27$.

However, the appearance of the specific symbols @ and #, while increasing their future reappearance probability, should not affect the estimate of the underlying distribution. This estimate should remain unchanged if any two other symbols appear instead, hence must depend only the sequence's *pattern* where the first two elements are identical, and the third element is different. It is the probability of this pattern, not of the observed values, that should be maximized.

Under the $(2/3, 1/3)$ maximum-likelihood distribution, the probability of this, first two same, third different, pattern is

$$\left(\frac{2}{3}\right)^2 \cdot \frac{1}{3} + \frac{2}{3} \cdot \left(\frac{1}{3}\right)^2 = \frac{2}{9}.$$

However, a $(1/2, 1/2)$ distribution is more likely to generate this pattern as it assigns it probability

$$\left(\frac{1}{2}\right)^2 \cdot \frac{1}{2} + \frac{1}{2} \cdot \left(\frac{1}{2}\right)^2 = \frac{1}{4}.$$

In Corollary 12 we show that the probability of this pattern is indeed maximized by the $(1/2, 1/2)$ distribution.

The difference between the maximum-likelihood distribution of @@# and the one proposed here is simple. Maximum likelihood postulates a distribution $(p, q)$, where $p$ is the probability of @ and $q$ is the probability of #, hence maximizes $p^2q$. The new distribution, by contrast, specifies only $(p, q)$, leaving open the question as to which of $p$ and $q$ is associated with @ and #, hence maximizes $p^2q + q^2p$.

Since we assume that the data sequences are sampled from an underlying distribution, the sequence of observations are independent and identically distributed (*i.i.d.*). Therefore, the order in which the elements are observed is irrelevant and the pattern probability is determined by the data's *profile*, the number of elements appearing any given number of times. For example, the sequences @@#, @#@, and #@@ share the same profile where one element appears once and one element appears twice, hence any underlying distribution will assign the same probability to their respective patterns. It follows that the distribution maximizing the pattern probability is determined by the data's profile, and we therefore call it the *high-profile* distribution.

In the previous example, the high-profile and maximum-likelihood distributions had different values but agreed on the size of the underlying support set. For some data samples, the support size differs too. Imagine that the first 20 outputs of an *i.i.d.* source consist of 10 distinct symbols, each appearing twice. Maximum likelihood would maximize the probability of the observed values, namely the product

$$\prod P^2(a)$$

taken over the 10 observed symbols, and would therefore hypothesize the uniform distribution over them. On the other hand, a uniform distribution over $k \geq 10$ symbols assigns to the observed pattern, and to all other patterns whose profile consists of 10 symbols, each appearing twice, a probability of

$$k^{\underline{10}}/k^{20},$$

where $k^{\underline{\ell}} \stackrel{\text{def}}{=} \frac{k!}{(k-\ell)!}$ is the $\ell$th *falling power* of $k$. Results in Section 4.7 imply that this probability is maximized by $k = 12$, and exceeds the probability assigned to the pattern by any other distribution.

Therefore, while maximum likelihood suggests that the distribution underlying this 20-element sequence has as many symbols as those observed, high profile proposes a distribution over a larger set. The latter better agrees with the intuition that in a sample whose size is comparable to the number of symbols in the universe, some fraction of the symbols will not appear.

The discrepancy between the high-profile and maximum-likelihood distributions grows as the number of observed symbols increases relative to the sample size. In the limit, consider again the $n$ DNA sequences. Maximum likelihood assigned probability $1/n$ to each previously observed sequence and 0 to all others. In better agreement with our intuition, high-profile would suggest that the sequences are generated by a distribution over a larger alphabet, ideally infinite, as such distributions assign the highest possible probability, 1, to patterns where each symbol appears once.

To quantify the difference between high-profile and maximum-likelihood distributions, consider predicting properties of unseen data. Suppose that having observed the $n$ DNA sequences, we try to estimate how many distinct sequences will be observed if we sample, say, $10n$ new individuals, and how many of these sequences will differ from those seen in the first sample. Maximum likelihood postulated a uniform distribution over the $n$ observed sequences. It would therefore suggest that the number of distinct sequences among the $10n$ new samples will be at most $n$, and that all the sequences will be among those that had appeared in the original sample. By contrast, high profile postulated a distribution over an infinite alphabet, hence would suggest that all sequences in the new sample will be distinct, and will differ from the original sequences. Clearly, the latter will better match the new observations.

The problems of identifying the distribution underlying a data sample and of using this information to predict properties of unseen data were considered by Fisher *et al.* in 1943 [Fisher *et al.*, 1943].

They assumed that the number of Malayan butterflies of each species was an independent Poisson random variable, and used this assumption to estimate the number of different species. In a different context, this problem was considered by Good and Turing in World War II, and further pursued by [Good, 1953, Good *et al.*, 1956], to yield a family of estimators that have since been used in a variety of applications *e.g.,* [Katz. 1987, Church *et al.*, 1991, Chen, 1996, Song, 1999].

Several other researchers have also considered the estimation of the probability of unseen data *e.g.,* [Robbins, 1968, Keener *et al.*, 1987, Mao *et al.*, 2002], and the problem of estimating the number of distinct elements in the underlying distribution [Efron *et al.*, 1976, Chao, 1984, Sichel, 1986, Zelterman, 1988]. Of these, [Efron *et al.*, 1976, Zelterman, 1988, Mao *et al.*, 2002] build on the model of [Fisher *et al.*, 1943]. Conditioned on the size of the sample, the Poisson samples become multinomial, and this was considered by [Chao, 1984, Sichel, 1986, Clayton *et al.*, 1987, Keener *et al.*, 1987]. See [Bunge *et al.*, 1993] for a overview of results obtained by the these researchers, and for other approaches to this problem.

In all these papers certain assumptions, such as a given prior, are made on the underlying distributions. By contrast, in this paper, we do not make any such assumptions, and yet the estimator we derive appears to address several of the same problems solved above, such as that of the "missing mass".

The high-profile estimate was motivated by work on the Good-Turing estimation problem [Good, 1953] and on universal compression over unknown alphabets.

Good-Turing estimation concerns estimation of the probability of the next element in a sequence after observing the previous ones when the collection of symbols is not known in advance. The estimator predicts whether the next symbol will be "new", or one of the symbols seen beforehand, and in the latter case, which one. It was shown in [Orlitsky *et al.*, 2003] that this problem is equivalent to predicting the pattern of the sequence, and an algorithm that does not asymptotically underestimate the probability of any sequence was derived.

Universal data compression concerns compression of the data when the underlying distribution is not known. Kieffer [Kieffer, 1978] showed that *i.i.d.* distributions over infinite alphabets cannot be universally compressed. Motivated by his result, most popular text compression algorithms, such as Lempel Ziv [Ziv *et al.*, 1977] and Context Tree Weighting [Willems *et al.*, 1995], operate on the smallest alphabet—bits rather than the more natural—words. In [Orlitsky *et al.*, 2003], compression of sequences was separated into two parts. Compression of the sequence's pattern, and of the correspondence between the indices and the individual symbols. It was shown that the pattern can be compressed with diminishing per-symbol redundancy. This shows that the reason for the infinite per-symbol redundancy shown by Kieffer is not the slow learning of the underlying distribution, but only the specification of the symbols appearing in the sequence.

## 2 SEQUENCES, PATTERNS, AND PROFILES

As mentioned in the introduction, we determine the probability of a sequence's pattern which, in turn, is determined by its profile. We now define these terms. Let $\overline{x} = x_1 \ldots x_n$ be a length-$n$ sequence. We call the distinct values appearing in $\overline{x}$, *symbols* and let $m = m(\overline{x})$ denote their number. The *index* $\imath(x)$ of a symbol $x$ is one more than the number of symbols preceding its first appearance in $\overline{x}$. The *pattern* of $\overline{x}$ is the index concatenation

$$\psi(\overline{x}) \stackrel{\text{def}}{=} \imath(x_1)\imath(x_2)\ldots\imath(x_n),$$

For example, the sequence $\overline{x}$ = "*abracadabra*" consists of $m = 5$ symbols, *a*,*b*,*r*,*c*, and *d*. Their indices are $\imath(a) = 1$, $\imath(b) = 2$, $\imath(r) = 3$, $\imath(c) = 4$, and $\imath(d) = 5$, hence

$$\psi(abracadabra) = 12314151231.$$

If a symbol $\psi$ in a pattern repeats $i$ times, we abbreviate it as $\psi^i$. For example, we write the pattern 11222113 as $1^2 2^3 1^2 3^1$. A pattern of the form $1^{\ _1} 2^{\ _2} \ldots m^{\ _m}$ where $\mu_1 \geq \mu_2 \geq \ldots \geq \mu_m$, is called *canonical.*

As mentioned earlier, the probability of a pattern is determined by its profile, defined in the rest of this section. The *multiplicity* of $x$ in $\overline{x}$ is

$$\mu_x \stackrel{\text{def}}{=} \mu_x(\overline{x}) \stackrel{\text{def}}{=} |\{i : x_i = x\}|,$$

the number of times $x$ appears in $\overline{x}$. The *minimum multiplicity* $\mu_{\min} \stackrel{\text{def}}{=} \mu_{\min}(\overline{x})$ of $\overline{x}$ is the smallest multiplicity of any of its elements, and the *maximum multiplicity* $\mu_{\max} \stackrel{\text{def}}{=} \mu_{\max}(\overline{x})$ of $\overline{x}$ is the largest multiplicity of any of its elements. The *prevalence* of an integer $\mu \in \mathbb{P}$ in $\overline{x}$ is

$$\varphi\ \ \stackrel{\text{def}}{=} \varphi\ \ (\overline{x}) \stackrel{\text{def}}{=} |\{x : \mu_x = \mu\}|,$$

the number of symbols appearing $\mu$ times in $\overline{x}$. The *profile* of $\overline{x}$ is the formal product

$$\bar{\varphi} \stackrel{\text{def}}{=} \bar{\varphi}(\overline{x}) \stackrel{\text{def}}{=} \prod_{=1}^{|\overline{x}|} \mu^{\varphi_\mu},$$

of all multiplicities and their prevalences, where $\mu^0$ terms are typically omitted.

For example, in the sequence *abracadabra*, the multiplicities are $\mu_a = 5$, $\mu_b = \mu_r = 2$, and $\mu_c = \mu_d = 1$, hence the minimum and maximum multiplicities are $\mu_{\min} = 1$ and $\mu_{\max} = 5$. The prevalences are therefore $\varphi_1 = 2$, $\varphi_2 = 2$, $\varphi_5 = 1$, and $\varphi_i = 0$ for all other $i$, and the profile is $\bar{\varphi} = 1^2 2^2 3^0 4^0 5^1 6^0 \cdot \ldots \cdot (11)^0 = 1^2 2^2 5^1$.

Patterns, as sequences, have profiles too. It is easy to see that the profile of a sequence is the same as that of its pattern. For example, the pattern of *abracadabra* is 12314151231 whose profile is also $1^2 2^2 5^1$. Given a profile $\bar{\varphi}$, we let

$$\bar{\Psi}_{\overline{\varphi}} \stackrel{\text{def}}{=} \{\bar{\psi} : \overline{\varphi}(\bar{\psi}) = \overline{\varphi}\}$$

be the set of patterns of profile $\bar{\varphi}$. For every $\overline{\varphi}$, exactly one of the patterns in $\bar{\Psi}_{\overline{\varphi}}$ is canonical, and is called the *canonical pattern* of $\overline{\varphi}$. For example, the canonical pattern of the profile $1^2 2^2 5^1$ is $1^5 2^2 3^2 4^1 5^1 =$ 11111223345.

## 3 THE HIGH-PROFILE DISTRIBUTION

We now define the distribution underlying the data and the probability it induces over sequences and patterns. Since the distribution maximizing the probability of a pattern may have an arbitrarily large support, it is mathematically convenient to model parts of a distribution as continuous.

A *discrete distribution* is a mapping $P : \mathcal{K} \to [0, 1]$ where $\mathcal{K}$ is a discrete set and $\sum_{a \in \mathcal{K}} P(a) = 1$. A *continuous distribution* is a mapping $P : \mathcal{X} \to [0, \infty)$ where $\mathcal{X}$ is an interval and $\int_{x \in \mathcal{X}} P(x)dx = 1$. A *mixed distribution* is a weighted combination of a discrete distribution over a set $\mathcal{K}$ and a continuous distribution over an interval $\mathcal{X}$ such that $\sum_{a \in \mathcal{K}} P(a) + \int_{\mathcal{X}} P(x)dx = 1$.

Suppose that $n \in \mathbb{P}$ data samples are collected sequentially from some distribution $P$. The observed samples will form an *i.i.d.* sequence $\overline{X} = X_1, X_2, \ldots, X_n$ where the probability of $\overline{x} = x_1, \ldots, x_n$ is

$$P(\overline{x}) \stackrel{\text{def}}{=} P(\overline{X} = \overline{x}) = \prod_{i=1}^{n} P(x_i).$$

The random sequence's pattern $\bar{\Psi} = \psi(\overline{X})$ will be distributed according to

$$P(\bar{\psi}) \stackrel{\text{def}}{=} P(\bar{\Psi} = \bar{\psi}) = P(\{\overline{x} : \psi(\overline{x}) = \bar{\psi}\}), \qquad (1)$$

the probability that a length-$n$ sequence generated according to $P$ has pattern $\bar{\psi}$.

For example, if $P$ is a discrete distribution over $\{a_1, a_2\}$ with $P(a_1) = p_1$ and $P(a_2) = p_2$, where $p_1 + p_2 = 1$, then for $n = 2$, the possible patterns are 11 and 12 and occur with probability

$$P(11) = P(\{a_1a_1, a_2a_2\}) = p_1^2 + p_2^2 \qquad \text{and}$$
$$P(12) = P(\{a_1a_2, a_2a_1\}) = 2p_1p_2.$$

By contrast, if $P$ is a continuous distribution over an interval $\mathcal{X}$ then, with probability one, two elements drawn according to $P$ will differ, hence

$$P(11) = P(\{xx : x \in X\}) = 0 \qquad \text{and}$$
$$P(12) = P(\{x_1x_2 : x_1 \neq x_2 \in X\}) = 1.$$

Finally, if $P$ is a mixed distribution over the union of $\{a_1, a_2\}$ and an interval $\mathcal{X}$, with $P(a_1) = p_1$, $P(a_2) = p_2$, and $P(\mathcal{X}) = q$, where $p_1 + p_2 + q = 1$, then

$$\begin{aligned} P(11) &= P(\{a_1a_1, a_2a_2\}) + P(\{xx : x \in X\}) \\ &= p_1^2 + p_2^2, \end{aligned}$$

and

$$\begin{aligned} P(12) &= P(\{a_1a_2, a_2a_1\}) + P(\{x_1x_2 : x_1 \neq x_2 \in \mathcal{X}\}) \\ &\quad + P(\{a_2x, xa_2 : x \in \mathcal{X}\}) + P(\{a_1x, xa_1 : x \in \mathcal{X}\}) \\ &= 2p_1p_2 + 2p_1q + 2p_2q + q^2 = 1 - p_1^2 - p_2^2. \end{aligned}$$

Note that the pattern probability is determined by the multiset $\{P(a) : a \in \mathcal{K}\}$ of discrete probabilities and the interval probability $P(\mathcal{X})$. The values of the elements in $\mathcal{K}$ and $\mathcal{X}$, the arrangement of the probabilities over $\mathcal{K}$, and the precise distribution over $\mathcal{X}$ do not matter. Furthermore, $P(\mathcal{X}) = 1 - \sum_{a \in \mathcal{K}} P(a)$, and is thus determined by the discrete probabilities. We therefore identify any distribution, discrete, mixed, or continuous, with a vector $P \stackrel{\text{def}}{=} (p_1, p_2, \ldots)$ in the *simplex*

$$\mathcal{P} \stackrel{\text{def}}{=} \{(p_1, p_2, \ldots) \ : \ p_i \geq 0 \text{ and } \sum_{i=1}^{\infty} p_i \leq 1\}.$$

If $P \stackrel{\text{def}}{=} (p_1, p_2, \ldots) \in \mathcal{P}$ is a distribution, then its *discrete probability* $p \stackrel{\text{def}}{=} \sum_i p_i$ is the total probability of all discrete elements, its *continuous probability* $q \stackrel{\text{def}}{=} 1 - p$ is the probability of all continuous elements, its *discrete support* $\mathcal{K} \stackrel{\text{def}}{=} \{i : p_i > 0\}$ is the set of discrete positive-probability elements, its *discrete size* is $k \stackrel{\text{def}}{=} |\mathcal{K}|$,the cardinality of the discrete support, and its *total size* $S$ is the cardinality of the set of all positive-probability elements, discrete or continuous. It equals $k$ if $P$ is discrete and is $\infty$ otherwise. Since the order of the probabilities $p_i$ does not affect the induced pattern probability, we assume that $p_i \geq p_{i+1}$, namely, $P = (p_1, p_2, \ldots)$ is *monotone*, and therefore lies in the *monotone simplex*

$$\mathcal{P}_{\mathcal{M}} \stackrel{\text{def}}{=} \{(p_1, p_2, \ldots) \ \in \mathcal{P} : p_i \geq p_{i+1}\}.$$

For example, the probability vector (1) represents a discrete distribution over a single element, the vector $(.3, .2)$ represents a mixed distribution over two elements with respective probabilities .3 and .2 and an interval with probability .5, and () represents a continuous distribution over an interval.

To determine the induced probabilities of general patterns, we need some definitions. Let $A$ and $B$ be sets. By analogy with the notation $B^A$ for functions from $A$ to $B$ and $\binom{A}{2}$ for 2-element subsets of A, let $B^{\underline{A}}$ denote the collection of all injections (one-to-one functions) from $A$ to $B$. Recall that $m = m(\bar{\psi})$, denotes the number of symbols in $\bar{\psi}$. Define $[m] = \{1, 2, \ldots m\}$. Then $f \in \mathbb{P}^{\underline{[m]}}$ maps the sequence $\bar{\psi} = \psi_1\psi_2 \ldots \psi_n \in [m]^n$ to the sequence $f(\bar{\psi}) \stackrel{\text{def}}{=} f(\psi_1)f(\psi_2) \ldots f(\psi_n) \in \mathbb{P}^n$ where $f(\psi_i) = f(\psi_j)$ iff $\psi_i = \psi_j$.

As before, let $\mu_j$ denote the number of times $j$ appears in $\bar{\psi}$. Equation (1) can therefore be expressed as

$$P(\bar{\psi}) = \sum_{I \subseteq \{i:\ \ _i=1\}} q^{|I|} \sum_{f \in \mathbb{P}^{\underline{[m]-I}}} \prod_{j \in [m]-I} P(f(j))^{\ j}. \tag{2}$$

Note that the number of nonzero products depends on the discrete size, but even if $k = \infty$, the sum converges to a unique value. Alternative forms of this equation, and simplification for finite $k$ exist, but are omitted from this abstract. As can be seen from this equation, for any probability $P$, the induced pattern probability $P(\bar{\psi})$ depends only on the pattern's profile $\overline{\varphi}(\bar{\psi})$. For example, $P(112) = P(121) = P(122)$.

We are interested in

$$\hat{P}(\bar{\psi}) \stackrel{\text{def}}{=} \sup_{P \in \mathcal{P}} P(\bar{\psi}), \tag{3}$$

the highest probability assigned to a pattern $\bar{\psi}$ by any distribution. As can be observed from (2), for any $\bar{\psi}$, $P(\bar{\psi})$ is a symmetric function in $P = (p_1, p_2, \ldots)$, which for finite $k$ is a degree-$n$ symmetric polynomial. We are therefore interested in the supremum of this function over all distributions $P$ in the simplex $\mathcal{P}$. Based on the previous observation, $\hat{P}(\bar{\psi})$ depends only on $\bar{\psi}$'s profile $\overline{\varphi}$, hence we call it the *high-profile probability* of $\overline{\varphi}$.

One of the first results we prove is that $\hat{P}(\bar{\psi})$ is always achieved by some distribution which we call the *high-profile distribution* of $\bar{\psi}$. We denote this distribution by $\hat{P} = \hat{P}_{\bar{\varphi}}$, its total size by $\hat{S} = \hat{S}_{\bar{\varphi}}$, its discrete size by $\hat{k} = \hat{k}_{\bar{\varphi}}$, and its continuous probability by $\hat{q} = \hat{q}_{\bar{\varphi}}$.

This terminology should be interpreted with some caution. The probability induced on the empty pattern and the pattern 1, is 1 for all underlying distributions, hence we call these two patterns and their profiles, *trivial*, and all other patterns and profiles *nontrivial*. For all nontrivial patterns addressed in this paper, $\hat{P}$ is unique. While this property may hold for all nontrivial patterns, we do not know whether that is indeed the case. Therefore all statements made in the paper apply to all possible high-profile distributions. For example, when we say that the high-profile support of a pattern is finite, that means that whether the pattern has one or more high-profile distributions, all have finite support.

Some high-profile distribution are easily found. *Constant* sequences, where a single symbol repeats, have pattern $1...1 = 1^n$ and profile $n^1$. Since any distribution over a single-element set always yields the constant sequence, $\hat{P}_{n^1} = (1)$, implying that $\hat{k}_{n^1} = \hat{S}_{n^1} = 1$, $\hat{q}_{n^1} = 0$, and $\hat{P}(\bar{\Psi} = 1^n) = 1$.

*All-distinct* sequences, where every symbol appears once, have pattern $12 \ldots n$ and profile $1^n$. If $P$ is a continuous distribution over an interval, then the probability that an *i.i.d.* sequence drawn according to $P$ consists of all distinct symbols is 1. Hence the high-profile distribution is the continuous distribution $\hat{P}_{1^n} = ()$, implying that $\hat{k}_{1^n} = 0$, $\hat{S}_{1^n} = \infty$, $\hat{q}_{1^n} = 1$, and $\hat{P}(\bar{\Psi} = 12 \ldots n) = 1$.

Unlike these simple examples, establishing the high-profile distributions of most other profiles, or some of their properties, seems hard and is the subject of this paper.

# 4 RESULTS

We prove several results for several quantities related to high-profile distributions: their total size, discrete size, continuous probability, and show that as the size of the sample increases relative to the number of distinct symbols, the high-profile and maximum-likelihood distributions coincide.

## 4.1 Existence of Maximum

We first show that the supremum in Equation (3) is achieved for all patterns.

**Theorem 1** For all patterns $\bar{\psi}$, there exists a dis-tribution $P \in \mathcal{P}$ achieving $\hat{P}(\bar{\psi})$.

**Proof outline** The set $\mathcal{P}$ is a subset of $\ell_2$, the space of infinite sequences $P = (p_1, p_2, \ldots)$ with

$$||P||_2 \stackrel{\text{def}}{=} \left(\sum_{i=1}^{\infty} p_i^2\right)^{\frac{1}{2}} < \infty.$$

To prove that the supremum is indeed achieved, we show that $\mathcal{P}$ along with the norm $||.||_2$ constitutes a compact space and that for every pattern $\bar{\psi}$, $P(\bar{\psi})$ is continuous over $\mathcal{P}$. Compactness is proved by showing that like $\ell_2$, the space $\mathcal{P}$ is complete namely, every Cauchy sequence in the space converges, and by proving that $\mathcal{P}$ is totally bounded namely, for all $\epsilon > 0$ there exists a finite number of open balls of radius $\epsilon$ that cover the space. The continuity of $P(\bar{\psi})$ for all patterns $\bar{\psi}$ is proved by induction on $m$. □

### 4.2 Continuous Probability

As demonstrated by the DNA example, the high-profile distribution may contain a continuous part. Its existence and probability are considered here. Clearly, if every symbol appears at least twice, namely if $\varphi_1 = 0$, then $\hat{P}$ must be discrete. We show that the same holds when exactly one of the symbols appears once.

**Theorem 2** For all patterns of length $\geq 2$ with $\varphi_1 \leq 1$, $\hat{P}$ is discrete.

**Proof outline** Given a mixed distribution $P$ with $q$ as the mass on its continuous part, we construct a discrete distribution $Q$ by adding a new element of mass $q$ to the discrete part of $P$. We then show that $Q(\bar{\psi}) > P(\bar{\psi})$. This shows that the $\hat{P}$ is discrete. □

A more refined question concerns the continuous probability $\hat{q}$. We show that the continous probability of the high-profile distribution is at most the fraction of elements appearing once.

**Theorem 3** For all non-trivial patterns

$$\hat{q} \leq \frac{\varphi_1}{n}.$$

**Proof outline** Let $\bar{\psi}$ be a non-trivial pattern where $\varphi_1$ symbols appear once and let $P = (p_1, p_2, \ldots)$ have a continuous probability $q \stackrel{\text{def}}{=} 1 - \sum_i p_i > \varphi_1/n$. We show that another distribution assigns to $\bar{\psi}$ a higher probability. For $0 \leq \alpha \leq 1$, define

$$Q^\alpha = \left(\frac{1-\alpha}{1-q}p_1, \frac{1-\alpha}{1-q}p_2, \ldots\right)$$

to be the distribution consisting of a continuous probability $\alpha$, and the suitably normalized discrete probabilities of $P$. For $0 \leq i \leq \varphi_1$ let $\psi_1^{n-i} \stackrel{\text{def}}{=} \psi_1 \ldots \psi_{n-i}$ denote the pattern obtained by omitting the last $i$ symbols of $\bar{\psi}$. Then we show that

$$Q^\alpha(\bar{\psi}) = \sum_{i=0}^{\varphi_1} \binom{\varphi_1}{i} (1-\alpha)^{n-i} \alpha^i Q^0(\psi_1^{n-i}).$$

Therefore the derivative of $Q^\alpha(\bar{\psi})$ is negative for all $\alpha > \varphi_1/n$ and since $q > \varphi_1/n$,

$$Q^{\varphi_1/n}(\bar{\psi}) > Q^q(\bar{\psi}) = P(\bar{\psi}).$$ □

### 4.3 Finitude of High-profile discrete Size

We prove that the discrete size $\hat{k}$ of the high-profile distribution is always finite. This result allows us to assume a concrete (finite) form of the high-profile distribution. To prove the result we show that the number $V(P) \stackrel{\text{def}}{=} |\{\hat{p}_1, \hat{p}_2, \ldots\}|$, of distinct probabilities in the high-profile distribution $\hat{P} = (\hat{p}_1, \hat{p}_2, \ldots)$ of a pattern is at most its length.

**Theorem 4** For any non-trivial pattern $\bar{\psi}$,

$$V(\hat{P}) \leq \min\{2^m, n-1\}.$$

**Proof outline** We show that all possible probability values in a high-profile distribution are roots of a low-degree polynomial and use that fact to upper bound their number. Let $\hat{P} = (\hat{p}_1, \hat{p}_2, \ldots)$ be the high-profile distribution of $\bar{\psi}$, the pattern in question.

The proof is based on the fact that for some $\lambda$ and all $j \geq 1$,

$$\left.\frac{\partial P(\bar{\psi})}{\partial p_j}\right|_{P=\hat{P}} = \lambda. \tag{4}$$

It can then be shown that

$$\frac{\partial P(\bar{\psi})}{\partial p_j} = Q(p_j) \stackrel{\text{def}}{=} \sum_{\mathcal{S}\subseteq[m]} c_{\mathcal{S}} (-1)^{|\mathcal{S}|-1} p_j^{\sum_{\ell\in\mathcal{S}} \ell-1}, \tag{5}$$

where $c_{\mathcal{S}}$ is a non-negative constant depending on $\mathcal{S}$.

By (4) and (5) all the distinct probability values of $\hat{P}$ are positive roots of $Q(x) - \lambda$. The theorem follows from Descartes' rule of signs which implies that the number of positive roots of $Q(x) - \lambda$ is at most $\min\{n-1, 2^m\}$. □

The theorem implies that the number of distinct probabilities in any high-profile distribution is finite. Since any probability can appear only a finite number of times we obtain the following.

**Corollary 5** The discrete size $\hat{k}$ of any high-profile distribution is finite. □

### 4.4 Upper Bound on $\hat{S}$

In the previous subsections we proved that $\hat{k}$ is finite and proved sufficient conditions for $\hat{k} = \hat{S}$. We now provide an upper bound for $\hat{S}$. Recall that $\mu_{\min}$, the minimum multiplicity, is the smallest number of times any symbol appears in $\bar{\psi}$.

**Theorem 6** For all non-trivial patterns,

$$\hat{S} \leq m + \frac{m-1}{2^{\ \min} - 2}. \qquad (6)$$

**Proof outline** If $\mu_{\min} = 1$ then the result holds trivially. Therefore we consider patterns with $\mu_{\min} \geq 2$. From Theorem 2, $\hat{S} = \hat{k}$. For notational simplicity let $k = \hat{k}$ and let $\hat{P} = (\hat{p}_1, \hat{p}_2, \ldots \hat{p}_k)$, with $\hat{p}_i \geq \hat{p}_{i+1}$ for all $1 \leq i \leq k$. We construct a new distribution $Q$ with probabilities $\hat{p}_i$, $1 \leq i \leq k-2$, and $\hat{p}_k + \hat{p}_{k-1}$ suitably ordered. We then show that

$$\hat{P}(\bar{\psi}) - Q(\bar{\psi}) \leq \sum_{i=1}^{m} \left( \frac{m-1}{k-m} - \sum_{a=1}^{i-1} \binom{\mu_i}{a} \right) \hat{p}_k{}^{i} C_i$$

where $C_i \geq 0$. If for all $i$

$$\frac{m-1}{k-m} - (2^{\ i} - 2) < 0,$$

then $Q(\bar{\psi}) > \hat{P}(\bar{\psi})$, a contradiction. Therefore there exists at least one $i$ for which

$$\frac{m-1}{k-m} \geq 2^{\ i} - 2.$$

Since $\mu_{\min}$ is the minimum multiplicity we have

$$k \leq m + \frac{m-1}{2^{\ \min} - 2}. \qquad \square$$

The theorem implies that if the number of times each symbol appears is at least the logarithm of the number of symbols, then the high-profile support size does not use any additional symbols.

**Corollary 7** For all non-trivial patterns such that $\mu_{\min} > \log(m+1)$, $\hat{k} = m$. $\square$

### 4.5 Lower Bound on $\hat{S}$

In the previous subsection we derived an upper bound on $\hat{S}$. We follow it up with a lower bound. Recall that $\mu_{\max}$, the maximum multiplicity, is the maximum number of times any symbol appears.

**Theorem 8** For all non-trivial patterns,

$$\hat{S} \geq m - 1 + \left( \frac{\sum_{j \in [m]} 2^{-\ j} - 2^{-\ \max}}{2^{\ \max} - 2} \right). \qquad (7)$$

**Proof outline** If $\mu_{\max} = 1$ then the pattern belongs to the class of all distinct patterns and $\hat{S}$ is infinite. Therefore we only need to consider the case when $\hat{k}$ is finite, say $k$. Let $\hat{P} = (\hat{p}_1, \hat{p}_2, \ldots, \hat{p}_k)$ with $\hat{p}_i \geq \hat{p}_{i+1}$ be the high profile distribution. We construct a new distribution $Q$ with probabilities $\hat{p}_i$, $2 \leq i \leq k$ and two elements with probability $\hat{p}_1/2$ each, suitably ordered. We then show that

$$\hat{P}(\bar{\psi}) - Q(\bar{\psi}) \leq \sum_{i=1}^{m} \left( \frac{\sum_{j \in [m]} 2^{-\ j} - 2^{-\ i}}{k-m+1} - 2^{\ i} + 2 \right) C_i,$$

where $C_i > 0$. If

$$\left( \frac{\sum_{j \in [m]-\{i\}} 2^{-\ j}}{k-m+1} - (2^{\ i} - 2) \right) < 0$$

for all $i$, then $Q(\bar{\psi}) > \hat{P}(\bar{\psi})$, a contradiction. Therefore

$$k \geq m - 1 + \min_i \left( \frac{\sum_{j \in [m]-\{i\}} 2^{-\ j}}{2^{\ i} - 2} \right).$$

We show that this minimum is achieved when $\mu_i = \mu_{\max}$. $\square$

This leads to the following Corollary.

**Corollary 9** For all non-trivial patterns if $\mu_{\max} < \log(\sqrt{m} + 1)$ then

$$\hat{S} > m.$$

Namely, if the number of times each symbol appears is at most half the logarithm of the number of symbols, the high-profile support size is larger than the number of symbols observed.

Considering profiles of the form $2^m$, where $m$ symbols repeat twice, we obtain from (6) that $\hat{k} \leq m + (m-1)/2 < \infty$, and from (7) that $\hat{k} \geq 1.125(m-1)$. Hence for arbitrarily large $m$, there are sequences whose high-profile support size is finite yet larger than $m$ by a constant factor. We will later determine $\hat{k}$ exactly.

### 4.6 High-profile Distribution of Binary Profiles

Starting with sequences consisting of just two symbols we build upon the results obtained for $\hat{S}$ to derive results on $\hat{P}$. Let $n_0$ and $n_1$ (wolog $1 \leq n_0 \leq n_1$) denote the number of times each of the two symbols appears in a sequence of length $n$. If both $n_0 \geq 2$ then it follows from Corollary 7 that $\hat{k} = 2$. We show that this is the case even when $n_0 = 1$ and $n_1 \geq 2$.

**Theorem 10** For any $n_1 \geq 2$, all patterns of the profile $1^1 {n_1}^1$ have $\hat{k} = 2$. $\square$

Combining Corollary 7 with Theorem 10 with a result by Alon [Alon, 1986], on the maxima of

$$p^{n_0}(1-p)^{n-n_0} + (1-p)^{n_0}p^{n-n_0},$$

we obtain the following theorem.

**Theorem 11** For all $1 \le n_0 \le n/2$,

$$\hat{P}_{(n-n_0)^1 n_0{}^1} = \begin{cases} (0.5, 0.5) & ((n-n_0)-n_0)^2 \le n, \\ \frac{1}{1+p}, \frac{p}{1+p} & ((n-n_0)-n_0)^2 > n, \end{cases}$$

and for all patterns of that profile

$$\hat{P}(\bar{\psi}) = \begin{cases} \frac{1}{2^{n-1}} & ((n-n_0)-n_0)^2 \le n, \\ \frac{p^{n_0}+p^{n-n_0}}{(1+p)^n} & ((n-n_0)-n_0)^2 > n, \end{cases}$$

where $p$ is the unique root in $(0,1)$ of the polynomial

$$n_0 \cdot p^{n-2n_0+1} - (n-n_0) \cdot p^{n-2n_0} + (n-n_0) \cdot p - n_0. \ \square$$

The $\hat{P}$ of several short patterns follows from this theorem.

**Corollary 12** All patterns with profile $1^1 2^1$, $1^1 3^1$, or $2^2$ have $\hat{P} = (1/2, 1/2)$. $\square$

### 4.7 Short and Simple Profiles

We use results hitherto obtained to determine $\hat{k}$ and $\hat{P}$ for several simple profiles. We begin with all profiles of length at most 4. Their high-profile support sizes and distributions are summarized in Table 1. Perhaps the most interesting of these profiles is $1^2 2^1$, where two symbols appear once and one symbol appears twice, as in the sequence @#$@ for example. This is the smallest profile for which the support size (5) is finite but exceeds the number of symbols appearing (3).

**Theorem 13** For all patterns of profile $\bar{\varphi} = 1^2 2^1$,

$$\hat{P} = (1/5, 1/5, 1/5, 1/5, 1/5) \quad \square$$

The *uniform profiles* are a class of interesting profiles. They are of the form $r^m$, namely each of the $m$ symbols appears $r$ times. We show that in general $\hat{P}$ is uniform over a finite alphabet larger than $m$.

**Theorem 14** For all $r$ and $m$,

$$\hat{P}_{r^m} = \left(\frac{1}{\hat{k}}, \dots, \frac{1}{\hat{k}}\right)$$

where

$$\hat{k} = \min\left\{k \ge m : \left(1+\frac{1}{k}\right)^{mr}\left(1-\frac{m}{k+1}\right) > 1\right\}.$$

Table 1: High-profile distribution of profiles of length at most 4.

| $\bar{\varphi}$ | Canonical $\bar{\psi}$ | $\hat{P}_{\bar{\varphi}}$ |
|---|---|---|
| $1^1$ | 1 | any distribution |
| $2^1$, $3^1$, $4^1$ | 11, 111, 1111 | (1) |
| $1^2$, $1^3$, $1^4$ | 12, 123, 1234 | () |
| $2^1 1^1$ | 112 | $(1/2, 1/2)$ |
| $3^1 1^1$ | 1112 | $(1/2, 1/2)$ |
| $2^2$ | 1122 | $(1/2, 1/2)$ |
| $2^1 1^2$ | 1123 | $(1/5, 1/5, 1/5, 1/5, 1/5)$ |

**Proof outline** From Theorem 2 we know that the high-profile support size is finite. Given $P$ distributed over $[k]$, we have

$$P(\bar{\psi}) = m! \sum_{\{s_1, s_2, \dots s_m\} \in \binom{[}{k}]m} \Pi_{i=1}^m P(s_i)^r.$$

Using Lagrange multipliers we obtain a set of polynomial equations that have to satisfied by the probabilities of $P$. We show that the number of solutions is limited by Descartes' rule of signs and that the only solution that can correspond to a maxima is the uniform distribution. The result then follows by maximizing over all uniform distributions. $\square$

When $m$ tends to infinity, the expression for $\hat{k}$ can be simplified.

**Corollary 15** For all patterns of profile $r^m$,

$$\lim_{m\to\infty} \frac{\hat{k}}{m} = \alpha$$

where $-\alpha \ln\left(1-\frac{1}{\alpha}\right) = r$. $\square$

### 4.8 Asymptotic Behavior of $\hat{P}$

In this section we prove that if the multiplicity of every symbol in a long sequence is proportional to the sequence length, then the high-profile and maximum-likelihood distributions are very close to each other. We then deduce that for any underlying distribution, as the sample size increases, the high-profile and maximum-likelihood distributions coincide.

Let $P^{\bar{\alpha}} = (\alpha_1, \alpha_2, \dots, \alpha_m)$ be a distribution over $[m]$ such that $\sum_{i=1}^m \alpha_i = 1$ and $\alpha_1 \ge \alpha_2 \ge \dots \ge \alpha_m$. Consider sequences in $[m]^n$ where each symbol $i$ appears $\alpha_i n$ times; for simplicity we assume that $\alpha_i n$ is an integer. Let $\varphi_{\bar{\alpha}}^n$ denote the sequence's profile. For example, if the $\alpha_i$'s are all distinct,

$$\varphi_{\bar{\alpha}}^n = \Pi_{i=1}^m (\alpha_i n)^1.$$

We show that as $n$ tends to infinity, $\hat{P}_{\varphi^n_{\bar{\alpha}}}$, the high-profile distribution of $\varphi^n_{\bar{\alpha}}$, tends to $P^{\bar{\alpha}}$, the maximum-likelihood estimate of the sequences.

**Theorem 16** As $n$ tends to infinity

$$\hat{P}_{\varphi^n_{\bar{\alpha}}} \to P^{\bar{\alpha}}$$

in terms of both the K-L divergence $D(P^{\bar{\alpha}}||\hat{P}_{\varphi^n_{\bar{\alpha}}})$ and the $\ell_1$ distance $||P^{\bar{\alpha}} - \hat{P}_{\varphi^n_{\bar{\alpha}}}||_1$.

**Proof outline** For any two distributions $P$ and $Q$, it is well known, *e.g.,* [Cover *et al.*, 1991], that

$$\frac{1}{2\ln 2}||P-Q||_1^2 \le D(P||Q).$$

Hence it suffices to prove the theorem for $D(P^{\bar{\alpha}}||\hat{P})$.

Let $\bar{\psi}$ be a pattern with profile $\varphi_{\bar{\alpha}}$. From Corollary 7 it follows that for sufficiently large $n$, $\hat{S} = m$. From (2), for any distribution $P = (p_1, \ldots, p_m)$ over $[m]$,

$$P(\bar{\psi}) = \sum_{f \in [m]^{\underline{[m]}}} P(f(\bar{\psi})).$$

For any $f \in [m]^{\underline{[m]}}$ and any sequence $\overline{x}$ of profile $\varphi_{\bar{\alpha}}$,

$$P(f(\overline{x})) = 2^{-nH(P^{\bar{\alpha}})} \cdot 2^{-nD\left(P^{\bar{\alpha}}||(p_{f(1)},\ldots,p_{f(m)})\right)},$$

where $H$ is the entropy function. Hence,

$$\hat{P}(\bar{\psi}) \ge P^{\bar{\alpha}}(\bar{\psi}) \ge 2^{-nH(P^{\bar{\alpha}})}.$$

Since $P^{\bar{\alpha}}$ is sorted it can be verified that for any permutation $f \in [m]^{\underline{[m]}}$,

$$D(P^{\bar{\alpha}}||\hat{P}) \le D\big(P^{\bar{\alpha}}||\big(\hat{p}_{f(1)}, \ldots, \hat{p}_{f(m)}\big)\big).$$

We use this fact to show that for all sufficeintly large $n$

$$D\ \ P^{\bar{\alpha}}||\hat{P}\ \ \le \frac{1}{n}\log m^{\underline{m}},$$

and the theorem follows. □

### 4.9 Computational Aspects

As indicated by the previous subsections, analytic calculation of the high-profile distribution of a given pattern appears to be difficult. Computational calculation seems prohibitive too and even when the distribution $P$ is given, evaluation of the pattern probability $P(\bar{\psi})$ according to Equation (2) requires exponentially many operations in the pattern's length.

Together with Sajama [Orlitsky *et al.*, 2004], we constructed an algorithm for computing the high-profile distribution of a pattern. The algorithm follows the Expectation Maximization (EM) methodology and finds a local maximum of the function $P(\bar{\psi})$ over $P \in \mathcal{P}$. To avoid the exponential calculation of the probabilities, the algorithm uses Markov chain sampling as in [Jerrum *et al.*, 1997].

The algorithm is capable of calculating the high-profile distribution of patterns of length over a thousand. Its output always lower bounds the high-profile probability, and for all profiles whose high-profile probability we could determine either analytically, or because the data was synthetically generated with sufficiently few symbols that the maximizing probability could be estimated, it came extremely close, both in terms of the underlying distribution $P$ and the resulting probability $P(\bar{\psi})$. Results on the performance of the algorithm on both synthesized and real data will be published in a followup paper [Orlitsky *et al.*, 2004].

### Acknowledgements

We thank Ian Abramson, Noga Alon, Ron Graham, Nikola Jevtić, and Sajama for helpful discussions.

## References

[Alon, 1986] N. Alon. On the number of certain subgraphs contained in graphs with a given number of edges. *Israel J. Math*, 53:97–120, 1986.

[Bunge *et al.*, 1993] J. Bunge and M. Fitzpatrick. Estimating the number of species: a review. *Journal of the American Statistical Association*, 88:364–373, 1993.

[Chao, 1984] A. Chao. Nonparametric estimation of the number of classes in a population. *Scandanavian Journal of Statistics: Theory and Applications*, 11:265–270, 1984.

[Chen, 1996] S.F. Chen and J. Goodman. An empirical study of smoothing techniques for language modeling. In *Proceedings of the Thirty-Fourth Annual Meeting of the Association for Computational Linguistics*, pages 310–318, San Francisco, 1996. Morgan Kaufmann Publishers.

[Church *et al.*, 1991] K.W. Church and W.A. Gale. Probability scoring for spelling correction. *Statistics and Computing*, 1:93–103, 1991.

[Clayton *et al.*, 1987] M.K. Clayton and E.W. Frees. Nonparametric estimation of the probability of discovering new species. *Journal of the American Statistical Association*, 82:305–311, 1987.

[Cover *et al.*, 1991] T. M. Cover and J. A. Thomas. *Elements of Information Theory.* Wiley-Interscience Publication, 1991.

[Efron *et al.*, 1976] B. Efron and R. Thisted. Estimating the number of unseen species: How many words did Shakespeare know. *Biometrika*, 63:435–447, 1976.

[Fisher *et al.*, 1943] R. Fisher, A. Corbet, and C. Williams. The relation between the number of species and the number of individuals in a random sample of an animal population. *Journal of Animal Ecology*, pages 42–48, 1943.

[Good, 1953] I. J. Good. The population frequencies of species and the estimation of population parameters. *Biometrika*, 40:237–264, Dec 1953.

[Good *et al.*, 1956] I. J. Good and G. H. Toulmin. The number of new species, and the increase in population coverage, when a sample is increased. *Biometrika*, 43(1/2):45–63, 1956.

[Jerrum *et al.*, 1997] M.R. Jerrum and A.J. Sinclair. *Approximation algorithms for NP-hard problems in The Markov chain Monte Carlo method: An approach to approximate counting and integration*, pages 482–520. PwS Publishing, Boston, 1997.

[Katz. 1987] S.M. Katz. Estimation of probabilities from sparse data for the language model component of a speech recognizer. *IEEE Transactions on Acoustics, Speech and Signal Processing*, ASSP-35(3):400–401, Mar 1987.

[Keener *et al.*, 1987] R. Keener, E. Rothman, and N. Starr. Distributions on partitions. *Annals of Statistics*, 15(4):1466–1481, 1987.

[Kieffer, 1978] J.C. Kieffer. A unified approach to weak universal source coding. *IEEE Transactions on Information Theory*, 24(6):674–682, Nov 1978.

[Mao *et al.*, 2002] C. Mao and B. Lindsay. A poisson model for the coverage problem with a genomic application. *Biometrika*, 89:669–682, 2002.

[Orlitsky *et al.*, 2003] A. Orlitsky, N.P. Santhanam, and J. Zhang. Universal compression of memoryless sources over unknown alphabets. To appear in IEEE Transactions of Information Theory.

[Orlitsky *et al.*, 2004] A. Orlitsky, Sajama, N.P. Santhanam, K. Viswanathan, and J. Zhang. Algorithms for modeling distributions over large alphabets. To appear 2004 IEEE Symposium on Information Theory.

[Robbins, 1968] H.E. Robbins. Estimating the total probability of the unobserved outcomes of an experiment. *Annals of Mathematical Statistics*, 39:256–257, 1968.

[Sichel, 1986] H.S. Sichel. The GIGP distribution model with applications to physics literature. *Czechoslovak Journal of Physics, Ser. B*, 36:133–137, 1986.

[Song, 1999] F. Song and W.B. Croft. A general language model for information retrieval (poster abstract). In *Research and Development in Information Retrieval*, pages 279–280, 1999.

[Willems *et al.*, 1995] F. M. J. Willems, Y. M. Shtarkov, and Tj. J. Tjalkens. The context-tree weighting method: basic properties. *IEEE Transactions on Information Theory*, 41(3):653–664, 1995.

[Zelterman, 1988] D. Zelterman. Robust estimation in truncated discrete distributions with application to capture-recapture experiments. *Journal of Statistical Planning and Inference*, 18:225–237, 1988.

[Ziv *et al.*, 1977] J. Ziv and A. Lempel. A universal algorithm for sequential data compression. *IEEE Transactions on Information Theory*, 23(3):337–343, 1977.